\documentclass{article}

\usepackage[preprint]{neurips_2026}

\usepackage[utf8]{inputenc}
\usepackage[T1]{fontenc}
\usepackage{hyperref}
\usepackage{url}
\usepackage{booktabs}
\usepackage{multirow}
\usepackage{amsfonts}
\usepackage{amssymb}
\usepackage{amsmath}
\usepackage{nicefrac}
\usepackage{microtype}
\usepackage{xcolor}
\usepackage{graphicx}
\usepackage{float}
\usepackage{placeins}
\usepackage{enumitem}
\newcommand{\ci}[2]{{\scriptsize\,(#1,\,#2)}}

\title{Enginuity: A Dataset and Benchmark for Vision-Language Understanding of Engineering Diagrams}

\author{%
  Abhishek Kumar \\
  Predii \\
  \texttt{abhishek.kumar@predii.com} \And
  Isha Motiyani \\
  Predii \\
  \texttt{isha.motiyani@predii.com} \And
  Tilak Kasturi \\
  Predii \\
  \texttt{tilak@predii.com} \AND
  Ethan Seefried \\
  Oak Ridge National Laboratory \\
  \texttt{seefriedej@ornl.gov} \And
  Prahitha Movva \\
  Independent Researcher \\
  \texttt{prahitha.movva03@gmail.com} \And
  Tirthankar Ghosal \\
  Oak Ridge National Laboratory \\
  \texttt{ghosalt@ornl.gov}
}

\begin{document}

\maketitle

\begin{abstract}
Engineering diagrams pose a distinct challenge for vision-language models: unlike natural images or general documents, they encode information through dense spatial layouts, domain-specific symbols, and cross-references between visual callouts and structured parts tables. Despite their centrality to service, repair, and design workflows, there is no public benchmark for measuring VLM capabilities in this domain; existing datasets primarily focus on flowcharts, scientific figures, or business documents. To address this gap, we introduce \textsc{Enginuity}, the first open dataset and benchmark for evaluating VLMs on complex engineering diagrams. We define two tasks over a corpus of U.S.\ military service and repair manuals: structured parts-table extraction (Task~1) and free-form visual diagram question answering (VQA)(Task~2) for benchmarking. We evaluate four frontier VLMs (GPT-5.2 Chat, Claude Opus~4.7, Gemma~4, Qwen3-VL-32B-Instruct) under zero-shot and chain-of-thought prompting. On Task~1, models reach Recall@all of 0.61--0.87 but Token~F1\textsubscript{pen} of only 0.03--0.18, exposing a systematic gap between part identification and description fidelity. Task~2 reveals a consistent factual-reasoning gap across all models. A supporting analysis shows that token-overlap metrics under-report model capability on technical descriptions by 2--6× relative to semantic similarity, motivating LLM-as-judge calibration for domain-specific evaluation. We release the dataset, annotations, evaluation harness, and per-sample model outputs to support a reproducible study of VLM capability on engineering content. 

\end{abstract}

\section{Introduction}
\label{sec:introduction}

Vision-language models have made rapid progress on document and chart understanding~\citep{mmmu, docvqa, chartqa}, yet \emph{engineering diagrams} (exploded parts views, electrical schematics, hydraulic circuits, and assembly drawings) remain comparatively under-evaluated. These diagrams encode dense, structured information through domain-specific conventions (leader lines, callouts, item-number indices, multi-column parts tables) that differ markedly from natural images or web charts. Engineering diagrams are produced and consumed at scale: automotive OEMs, defense suppliers, and industrial operators rely on them across the full product lifecycle. A VLM that could reliably parse such diagrams would augment technicians, accelerate training-material authoring, and enable structured extraction from legacy documentation.

\noindent\textbf{Why engineering diagrams are hard.}
Engineering diagrams differ from natural images in three ways that stress current VLMs. First, the key signal is often a small numeric callout embedded in a dense mechanical rendering; grounding it to the correct part in an adjacent table requires fine-grained spatial reasoning. Second, the surrounding text uses domain-specific codes (NSN, CAGEC, SMR) and alphanumeric part numbers absent from general web pretraining. Third, diagram and parts table form a tightly coupled pair, and neither conveys complete information without the other. Recent work finds frontier VLMs fall below 60\% accuracy on structured scientific diagrams~\citep{seephys}; engineering diagrams are strictly harder because domain vocabulary is scarce and correct answers require grounding spatial layout against a relational parts schema.

\noindent\textbf{Why a new dataset and benchmark.}
Existing VLM benchmarks~\citep{mmmu, docvqa, chartqa, ai2d, scienceqa} do not cover engineering content at meaningful scale. To our knowledge, no dataset pairs complex engineering diagrams from technical service manuals with structured parts-table ground truth at a scale suitable for evaluating frontier VLMs across downstream tasks such as structured parts extraction, diagram-grounded question answering, and multimodal information retrieval. \textsc{Enginuity} fills this gap. Diagrams are formally linked to manufacturer parts tables, requiring models to resolve item numbers and descriptions across visual and tabular modalities. The corpus spans 18 vehicle subsystems and 6 diagram types, reflecting real-world documentation breadth. Task-1 ground truth is derived directly from the manuals' own parts tables, not crowdsourced or synthetically generated. The closest industrial benchmark targets machinery photographs~\citep{industryeqa}; prior task-specific parsers~\citep{autodigi, autoengg} require custom pipelines that do not generalise to frontier VLMs.

\noindent\textbf{Building on prior work.}
Seefried et al.~\citep{enginuity} proposed the Enginuity concept: an open, large-scale, multi-domain engineering diagram dataset with structural annotations, targeting 50K diagrams and a CVPR~2026 shared task. That work is a vision paper---no data were released, no tasks were defined, and no evaluation was conducted. The present paper delivers on that proposal: 2{,}056 diagram--parts-table pairs with task-aligned extraction ground truth, 60 domain-expert-authored Q\&A pairs, and a systematic zero-shot evaluation of four frontier VLMs across two tasks---none of which exist in the prior release.

\noindent\textbf{Contributions.}
\begin{itemize}[nosep]
  \item \textbf{A benchmark dataset for engineering-diagram understanding.} \textsc{Enginuity} comprises 2,056 diagram--parts-table pairs from U.S.\ military service manuals; 494 figures are benchmarked on Task~1 (structured parts extraction) and 60 expert-authored Q\&A pairs form Task~2.
  \item \textbf{Domain-expert Q\&A ground truth.} Task-1 labels come directly from the manuals' own parts tables; Task-2 pairs are authored by a domain expert covering factual lookup and multi-step reasoning.
  \item \textbf{Benchmarking with Frontier LLMs.} We evaluate GPT-5.2 Chat, Claude Opus~4.7, Gemma~4, and Qwen3-VL-32B on structured parts extraction (Task~1) and free-form diagram Q\&A (Task~2) under zero-shot and chain-of-thought conditions.
\end{itemize}
\section{Related Work}
\label{sec:related_work}

\noindent\textbf{Vision-language model benchmarks.}
General-purpose multimodal benchmarks such as MMMU~\citep{mmmu} and MMBench~\citep{mmbench} cover a small proportion of engineering content but not at meaningful scale from service documentation. Document and chart benchmarks (DocVQA~\citep{docvqa}, ChartQA~\citep{chartqa}, InfographicsVQA~\citep{infographicvqa}, and DUDE~\citep{dude}) probe structured extraction from visually complex layouts, but draw content from business documents and web charts rather than technical manuals where the key signal is a numeric callout cross-referenced against a domain-specific parts schema.

\noindent\textbf{Document understanding.}
Layout-aware models such as LayoutLMv3~\citep{layoutlmv3} and document-foundation models such as Pix2Struct~\citep{pix2struct} advance structured extraction from forms and reports, but assume a grid of text fields, an assumption that breaks down for engineering diagrams where ground truth requires aligning visual callouts in a rendered image with rows in a separately typeset parts table.

\noindent\textbf{Diagram understanding.}
Prior work on diagram understanding has focused on flowcharts~\citep{flowvqa}, scientific figures, and textbook illustrations~\citep{ai2d,scienceqa}. SeePhys~\citep{seephys} shows frontier VLMs achieve sub-60\% accuracy on structured scientific diagrams by exploiting textual labels rather than visual structure, a failure mode we expect to be more pronounced on industrial engineering diagrams where domain vocabulary is absent from pretraining. ChartMuseum~\citep{chartmuseum} further demonstrates that LLM-judge scores can mask perceptual reasoning failures, motivating our dual evaluation protocol for Task~2.

\noindent\textbf{Engineering-domain AI.}
Prior work on engineering diagram digitization has produced task-specific pipelines combining symbol detection and graph search~\citep{autodigi, autoengg}; a recent survey covers the broader landscape~\citep{reviewdl}. These systems achieve strong results on specific diagram types but do not generalise across conventions or subsystems. IndustryEQA~\citep{industryeqa}, the closest existing benchmark to Enginuity, covers industrial machinery photographs rather than structured technical schematics and does not require table extraction against a ground-truth parts schema.
\section{The \textsc{Enginuity} Dataset}
\label{sec:dataset}
\textsc{Enginuity} is constructed from a corpus of U.S.\ military service and repair manuals spanning ground vehicles, aircraft engines, and heavy equipment. It comprises two components: a Task-1 dataset of diagram--parts-table pairs assembled through an automated extraction pipeline, and a Task-2 dataset of question--answer pairs authored by a domain-expert annotator. This section describes the source corpus, construction process, and resulting dataset characteristics; tasks are defined formally in Section~\ref{sec:tasks}.

\begin{figure}[t]
  \centering
  \includegraphics[width=\textwidth]{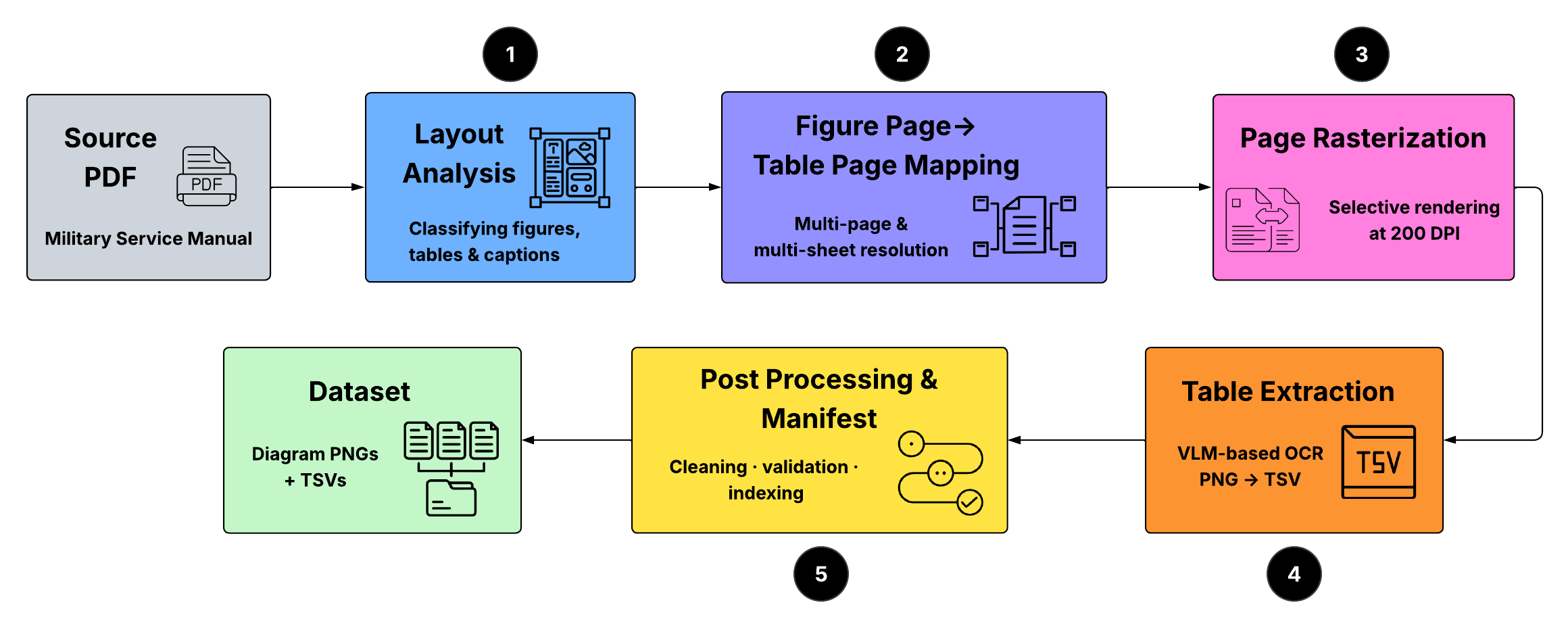}
  \vspace{-4pt}
  \caption{The \textsc{Enginuity} Task-1 construction pipeline. Five automated stages transform raw service manual PDFs into structured ground-truth parts-table TSVs linked to their corresponding diagram images.}
  \vspace{-6pt}
  \label{fig:pipeline}
\end{figure}

\subsection{Source Corpus}
\label{sec:dataset:sources}
The source corpus consists of 10 U.S. military service and repair manuals, totaling 6,940 pages. Four properties motivated the selection: (i)~every assembly diagram is paired with a formally structured parts table, making ground truth derivable without additional annotation; (ii)~the manuals span diverse vehicle subsystems and diagram types; (iii)~they are publicly available as government documentation; and (iv)~the engineering content is genuinely complex (multi-page diagrams, dense tabular data, and domain-specific part descriptions), making \textsc{Enginuity} a meaningful stress test for frontier VLMs.

\begin{figure*}[t]
  \vspace{-8pt}
  \centering
  \setlength{\tabcolsep}{3pt}
  \renewcommand{\arraystretch}{0.8}
  \begin{tabular}{ccc}
\includegraphics[width=0.325\textwidth,height=5cm,keepaspectratio]{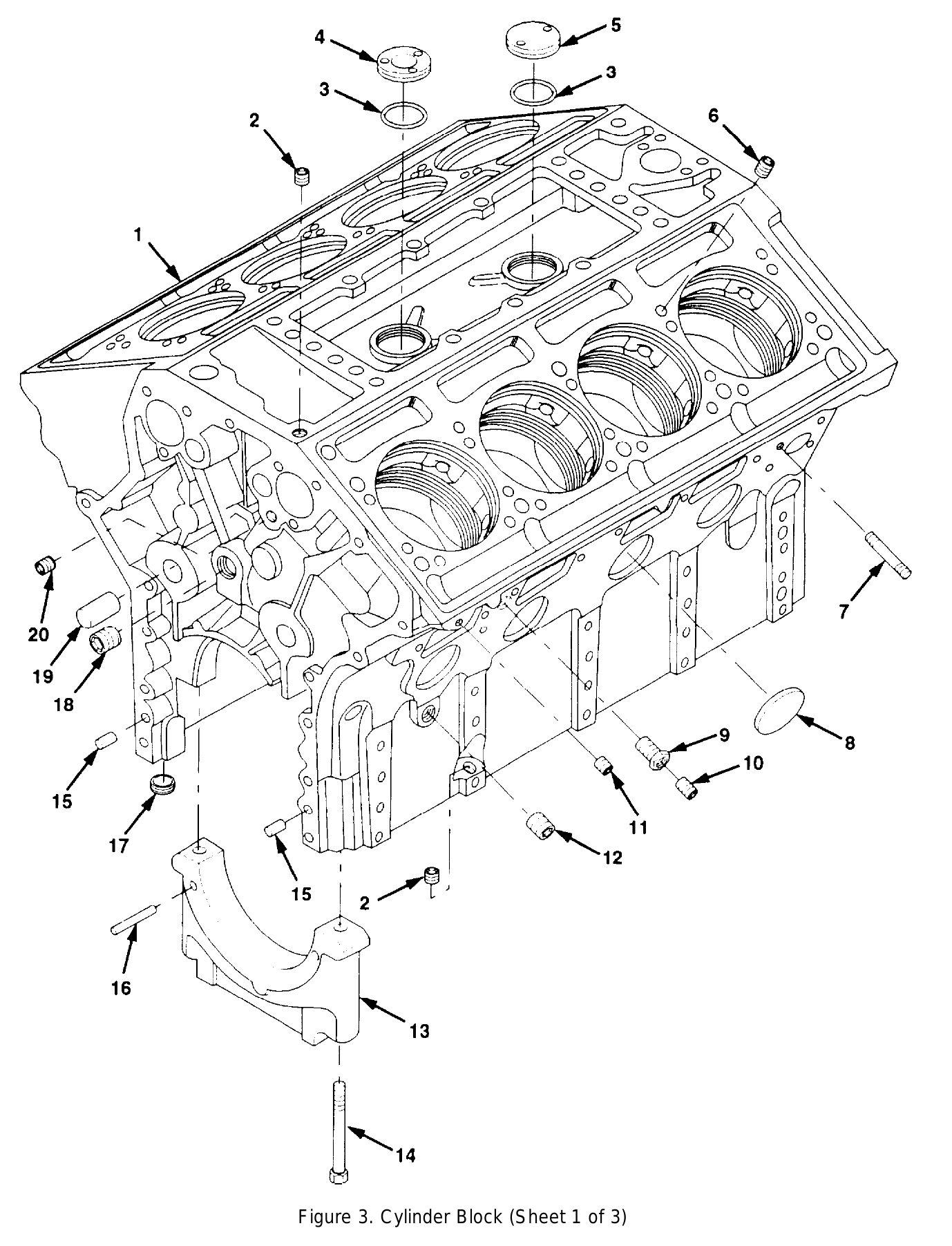} &
    \includegraphics[width=0.325\textwidth,height=5cm,keepaspectratio]{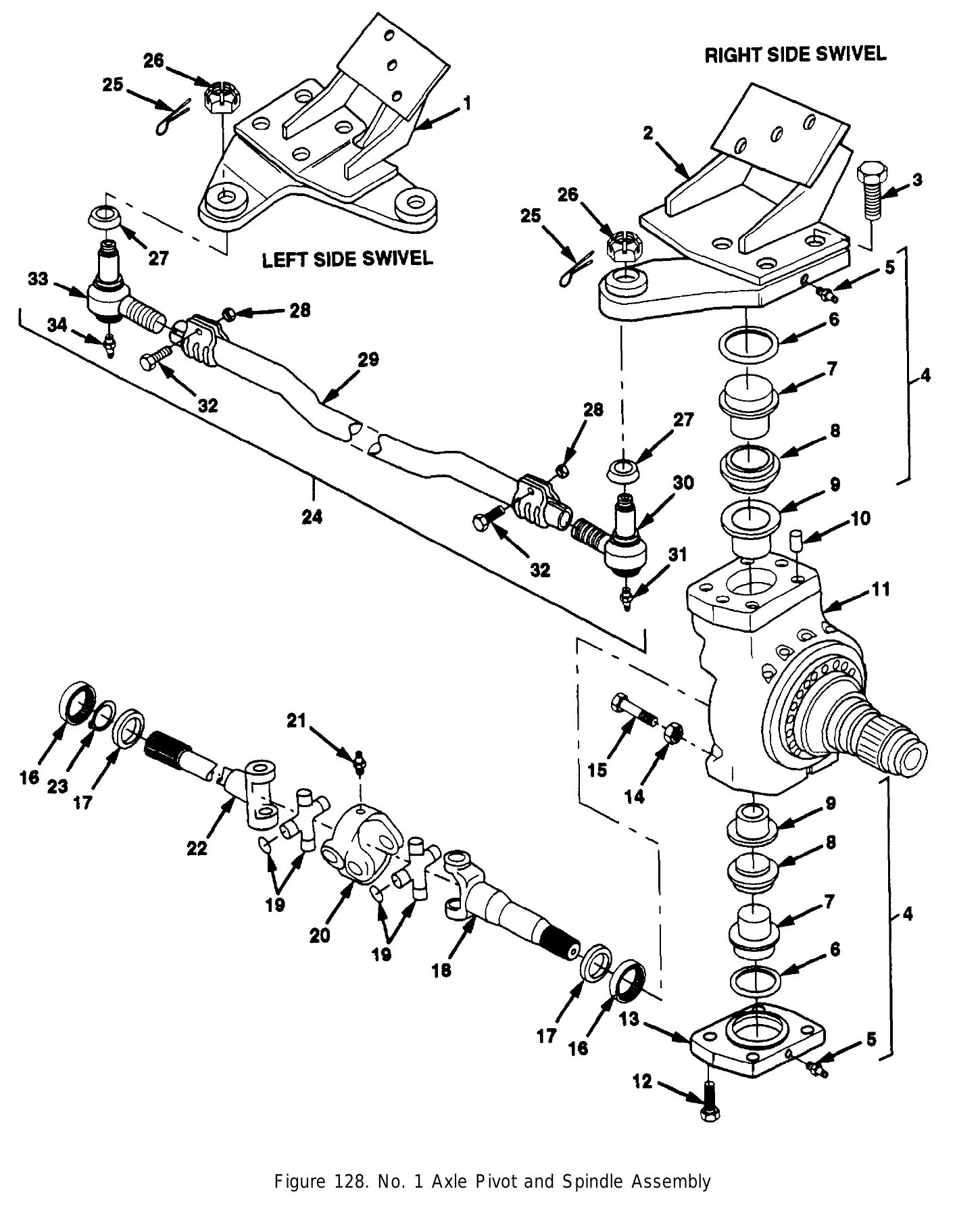} &
    \includegraphics[width=0.325\textwidth,height=5cm,keepaspectratio]{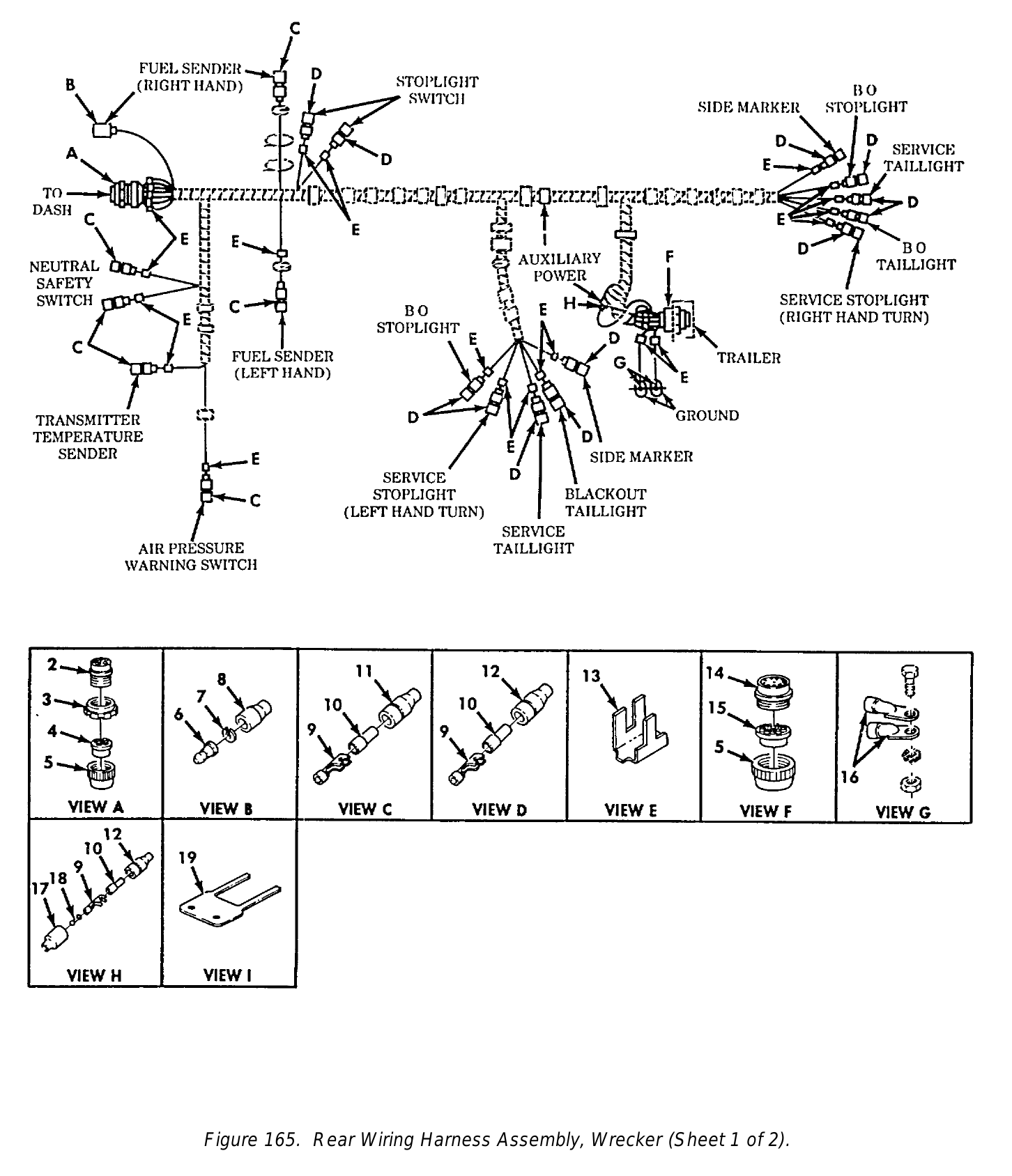} \\[2pt]
    \small\textbf{Parts / Assembly Diagram} &
    \small\textbf{Assembly / Exploded Parts View} &
    \small\textbf{Wiring / Electrical Diagram} \\[2pt]
    \includegraphics[width=0.325\textwidth,height=5cm,keepaspectratio]{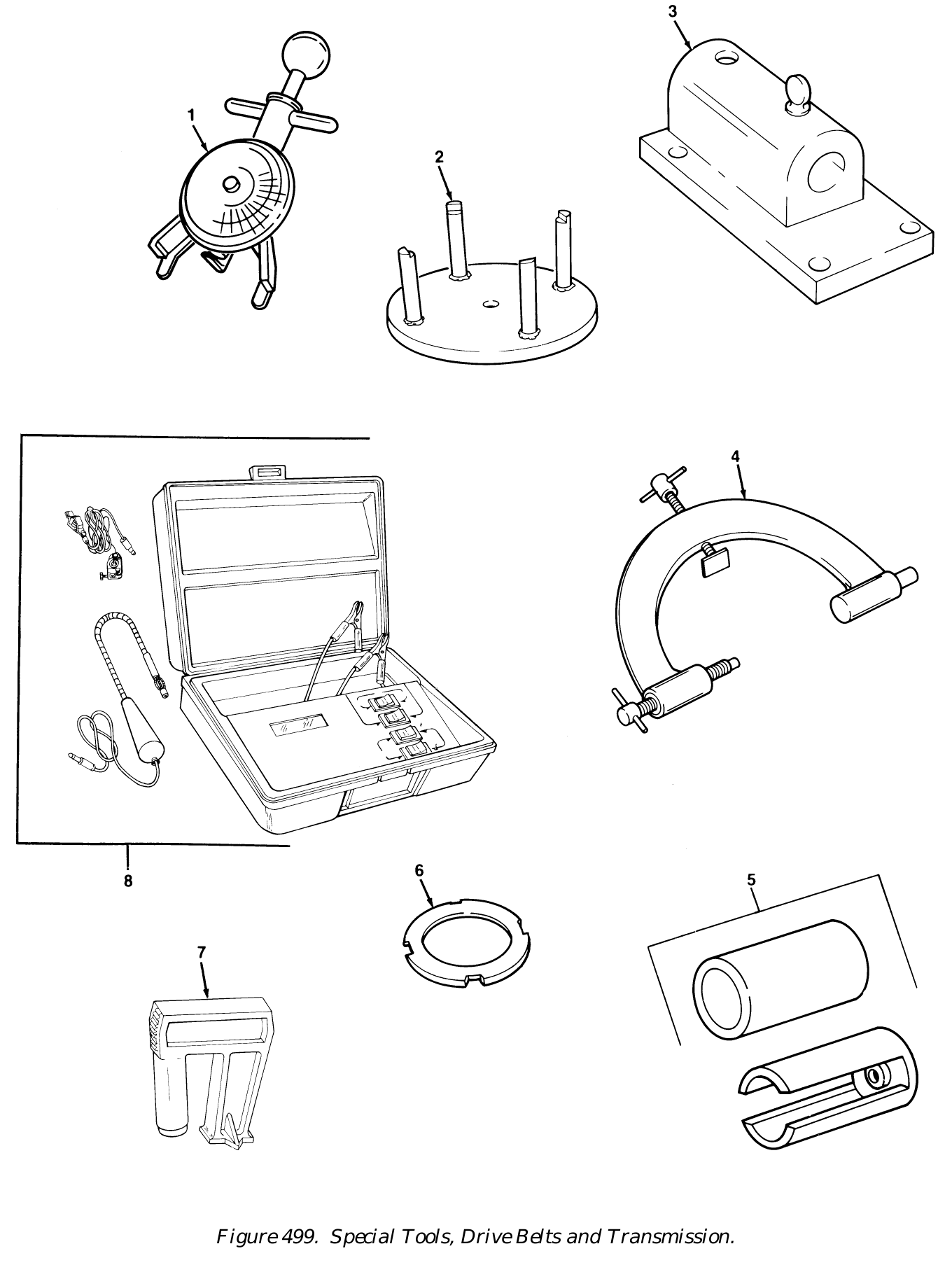} &
    \includegraphics[width=0.325\textwidth,height=5cm,keepaspectratio]{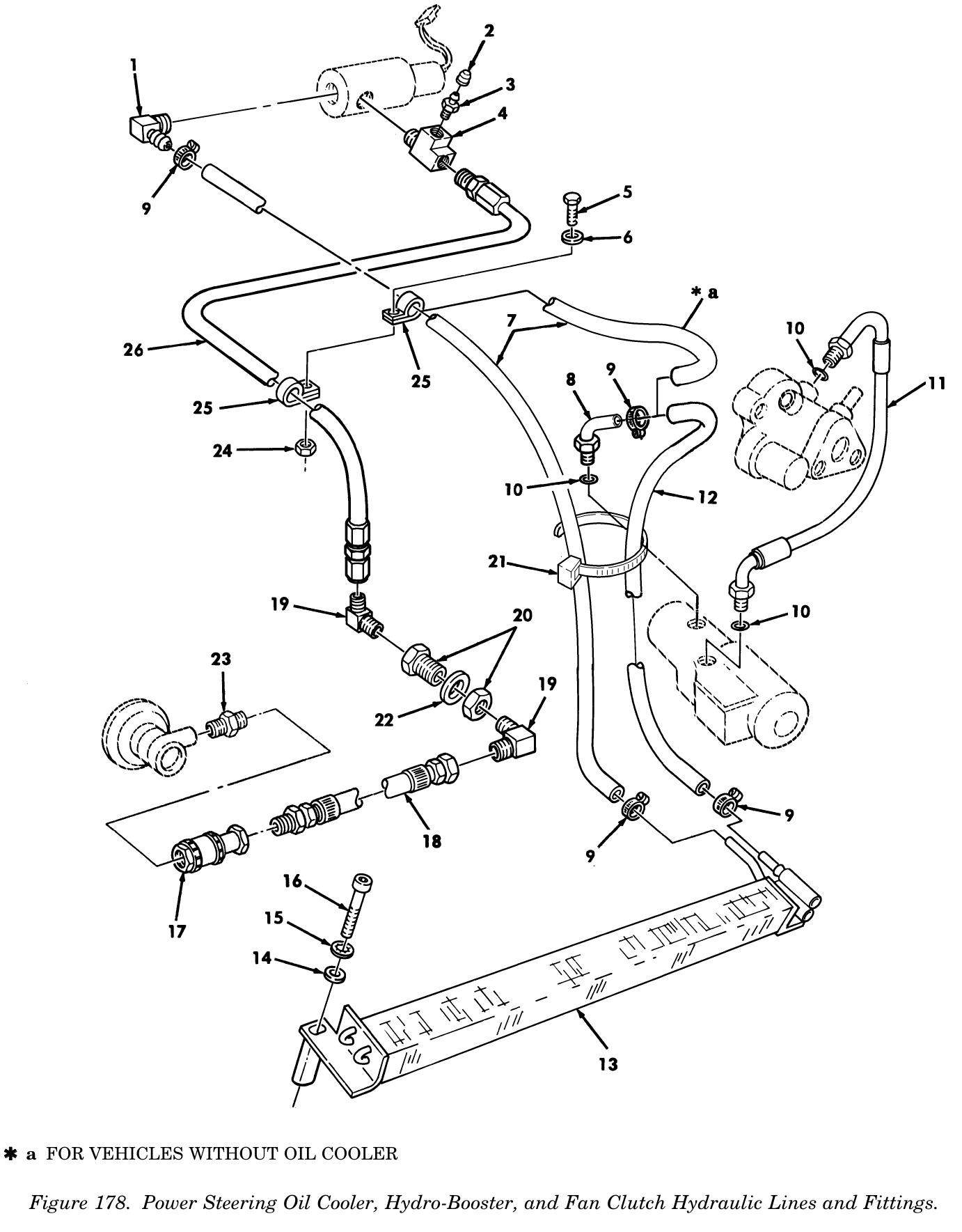} &
    \includegraphics[width=0.325\textwidth,height=5cm,keepaspectratio]{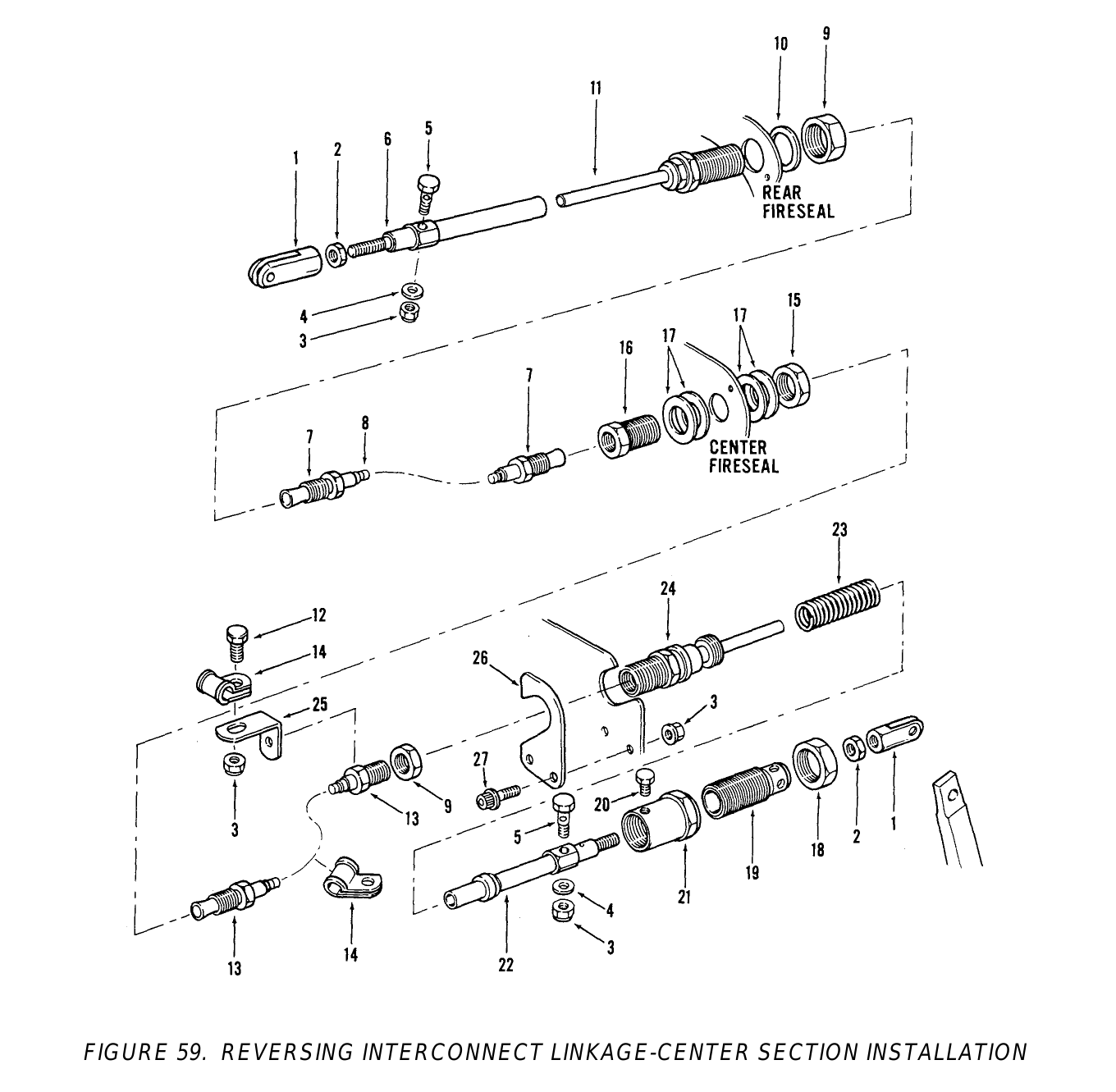} \\[2pt]
    \small\textbf{Equipment \& Tools Diagram} &
    \small\textbf{Hydraulic / Fluid Diagram} &
    \small\textbf{Cross-Section View} \\
  \end{tabular}
  \vspace{-4pt}
  \caption{Representative examples of the six diagram types in \textsc{Enginuity}, each requiring qualitatively different visual reasoning capabilities.}
  \vspace{-6pt}
  \label{fig:taxonomy-examples}
\end{figure*}
\subsection{Dataset Creation}
\label{sec:dataset:creation}
\subsubsection{Task-1: Structured Parts Extraction}
\label{sec:dataset:t1}

The Task-1 dataset consists of 2,056 diagram--parts-table pairs, each comprising a diagram image, one or more parts table images, and a structured ground-truth TSV with five fields: \texttt{item\_no}, \texttt{part\_no}, \texttt{description}, \texttt{uoc}, and \texttt{quantity} (Table~\ref{tab:dataset-stats}). The global manifest records 3,011 diagram--table pairings, since a single diagram may reference multiple table pages covering different sub-assemblies. Figure~\ref{fig:pipeline} illustrates the five-stage construction pipeline; the pipeline code is available at \url{https://github.com/abhishek-predii/engineering-diagrams-parsing}.

\noindent\textbf{Stage 1: Document layout analysis.}
Each source PDF is processed by the \texttt{Unstructured} library in \texttt{hi\_res} mode, which invokes a layout-detection model to classify every element on every page (image blocks, table regions, text paragraphs, and figure captions), recording each element's page number, bounding box, and type in a per-document metadata file.

\noindent\textbf{Stage 2: Figure--table page mapping.}
The metadata is parsed to match each figure caption to the sequence of table pages immediately following it in document order. Two structural patterns required special handling: diagrams spanning multiple table pages (up to five) are linked to the same figure entry; multi-sheet assemblies (e.g.\ ``Sheet~1 of~2'') are registered as distinct entries, each pointing to the same table pages but representing a different sub-assembly view.

\noindent\textbf{Stage 3: Page rasterization.}
Only the pages identified in Stage~2 are rendered, avoiding the cost of processing entire manuals. Page numbers are deduplicated across figure and table sets before rasterization, ensuring each page is rendered exactly once regardless of how many figures reference it. Matched pages are rasterized to PNG at 200\,DPI using \texttt{pdf2image} with the Poppler backend, preserving fine-detail line art and small numeric callouts. All rasterized figure images were manually reviewed to confirm each page contained a genuine engineering diagram.

\noindent\textbf{Stage 4: Table extraction via Claude Sonnet 4.5.}
Each table page PNG is submitted to Claude Sonnet~4.5~\citep{claude_sonnet_45} with a structured prompt requesting the five-column schema as a TSV. We chose a frontier VLM over conventional OCR because the content is visually complex: columns are not always cleanly separated, domain abbreviations require contextual interpretation, and multi-line cells and multi-level headers must be flattened into the target schema. Preliminary experiments showed conventional OCR produced unacceptably high error rates. Ground-truth quality was validated by human annotators who scored a stratified sample of 50 TSVs against their source table images on item number and description, yielding 98.7\% item extraction accuracy (819/830 items; 43 of 50 figures extracted without a single error).

\noindent\textbf{Stage 5: Post-processing, validation, and manifest construction.}
Each TSV undergoes structural cleaning: near-empty columns are dropped and adjacent duplicate columns are merged. A TSV is accepted only if it retains both \texttt{item\_no} and \texttt{description} columns with at least one data row, excluding non-parts-list pages such as wiring diagrams. A final filter removes figures linked to more than five table pages, discarding anomalous pairings from layout-parser errors. Per-PDF manifests are merged into a global manifest recording the diagram path, table path, TSV path, figure number, and source manual identifier.

\subsubsection{Task-2: Diagram Question Answering}
\label{sec:dataset:t2}

The Task-2 dataset comprises 12 diagrams and 60 question--answer pairs authored by a domain expert (Table~\ref{tab:dataset-stats}). The diagrams are selected from the Task-1 corpus to span 11 vehicle subsystems: engine block, cylinder head, lubrication, fuel, turbocharger, cooling, electrical, transmission, transfer case, brakes, and steering. Each diagram is paired with its corresponding parts table, and the annotator worked from both simultaneously to reflect the cross-modal nature of real-world service queries.

Each diagram receives five free-form question--answer pairs. Questions span two broad types: \emph{factual lookup} (directly identifying a part, item number, quantity, or part number from the table) and \emph{applied reasoning} (synthesizing information across diagram and table to answer service and maintenance questions, such as identifying required replacement hardware during an assembly procedure, determining component inclusion in a kit, or reasoning about system behavior). All pairs are reviewed for answerability and grounding in the source material before inclusion.
\subsection{Taxonomy of Engineering Diagrams}
\label{sec:dataset:taxonomy}
We characterise the \textsc{Enginuity} corpus along three axes: \emph{diagram type}, \emph{vehicle subsystem}, and \emph{complexity}. Diagram types and subsystems are identified by keyword matching on figure titles extracted during layout analysis; complexity is measured by label count, the total number of distinct part entries in the associated parts-table TSV(s) for a given diagram.

\noindent\textbf{Diagram types.}
Six diagram types were identified across the 2,056-figure corpus (Table~\ref{tab:diagram-types}). Parts and assembly diagrams account for nearly 90\% of the corpus; the remainder each require qualitatively different visual reasoning capabilities.

\noindent\textbf{Vehicle subsystems.}
The corpus spans 18 vehicle subsystems, ranging from powertrain (engine, transmission, drivetrain, fuel system) and chassis (suspension, steering, brakes, frame) to electrical systems, hydraulic and fluid lines, HVAC, fire suppression, armour protection, medical equipment, and communications. This breadth exposes the benchmark to highly specialised domain vocabulary across all subsystem areas.

\noindent\textbf{Complexity distribution.}
Label counts have a mean of 23 and a median of 17; full statistics are in Appendix~\ref{app:complexity}.

\begin{table*}[!t]
  \centering
  \begin{minipage}[t]{0.48\linewidth}
    \centering
    \caption{Diagram type distribution across the Task-1 corpus (2,056 figures).}
    \label{tab:diagram-types}
    \begin{tabular}{lr}
      \toprule
      \textbf{Diagram Type} & \textbf{Figures} \\
      \midrule
      Parts / Assembly Diagram          & 1{,}235 \\
      Assembly / Exploded Parts View    &   586   \\
      Equipment \& Tools Diagram        &    98   \\
      Wiring / Electrical Diagram       &    93   \\
      Hydraulic / Fluid Diagram         &    26   \\
      Cross-Section View                &    18   \\
      \midrule
      \textbf{Total}                    & \textbf{2{,}056} \\
      \bottomrule
    \end{tabular}
  \end{minipage}%
  \hspace{0.04\linewidth}%
  \begin{minipage}[t]{0.48\linewidth}
    \centering
    \caption{Dataset statistics for \textsc{Enginuity}.}
    \label{tab:dataset-stats}
    \begin{tabular}{llr}
      \toprule
      \textbf{Task} & \textbf{Quantity} & \textbf{Value} \\
      \midrule
      \multirow{7}{*}{\textit{Task-1}} & Source manuals             & 10 \\
                                       & Total source pages         & 6{,}940 \\
                                       & Unique diagrams            & 2{,}056 \\
                                       & Figure--table pairings     & 3{,}011 \\
                                       & Diagram types              & 6 \\
                                       & Vehicle subsystems         & 18 \\
                                       & Benchmark subset (figures) & 494 \\
      \midrule
      \multirow{3}{*}{\textit{Task-2}} & Diagrams                   & 12 \\
                                       & Q\&A pairs                 & 60 \\
                                       & Subsystems covered         & 11 \\
      \bottomrule
    \end{tabular}
  \end{minipage}
\end{table*}

\section{Tasks}
\label{sec:tasks}

\textsc{Enginuity} defines two tasks targeting complementary aspects of engineering-diagram understanding: structured component identification (Task~1) and free-form diagram question answering (Task~2).

\subsection{Task 1: Component Identification}
\label{sec:tasks:t1}

\noindent\textbf{Definition.}
Given an exploded parts diagram, the model must produce a structured table listing every called-out component, with each row specifying: \emph{item number} (the numeric callout in the diagram), \emph{part number}, \emph{description} (free-text technical description), and \emph{quantity}.

\noindent\textbf{Input.} A single rasterized diagram image.

\noindent\textbf{Output.}
A JSON array of objects with keys \texttt{item\_no}, \texttt{part\_number}, \texttt{description}, and \texttt{quantity}. We constrain output via prompt formatting and parse with a tolerant JSON reader to avoid penalizing minor formatting deviations (see Appendix~\ref{app:prompts}). \texttt{item\_no} is used solely for matching predicted rows to ground-truth rows; \texttt{description} is the only field scored.

\noindent\textbf{Metrics.}
We report:
\begin{itemize}[nosep]
  \item \textbf{Recall@all} --- fraction of ground-truth items recovered, matched on item number.
  \item \textbf{Token F1\textsubscript{pen} (description)} --- token-overlap F1 between predicted and ground-truth descriptions for matched items, penalised by the fraction of missed ground-truth parts (i.e.\ multiplied by Recall@all).
  \item \textbf{Semantic Similarity\textsubscript{pen}} --- cosine similarity between sentence embeddings of matched predicted and ground-truth descriptions, penalised for missed parts in the same manner as Token F1\textsubscript{pen}. Embeddings are computed with \texttt{all-MiniLM-L6-v2} via the \texttt{sentence-transformers} library~\citep{reimers-2019-sentence-bert}.
\end{itemize}

\subsection{Task 2: Diagram Question Answering}
\label{sec:tasks:t2}

At 60 expert-authored questions spanning 12 diagrams, Task~2 functions as a \emph{diagnostic probe} of open-ended diagram understanding rather than a statistically powered large-scale benchmark; results should be interpreted as initial signal on model behaviour on this class of queries.

\noindent\textbf{Definition.}
Given an engineering diagram page and a natural-language question, the model produces a free-form answer. Questions span direct lookup (e.g., ``What is the part number of item~14?'') to multi-step reasoning requiring spatial and tabular interpretation.

\noindent\textbf{Input.}
Each input is a composite image formed by vertically concatenating the primary exploded diagram with all associated parts-table images from the same manual page, ensuring questions requiring table lookups can be answered from the provided image.

\noindent\textbf{Output.}
A free-form natural-language answer to the question.

\noindent\textbf{Metrics.}
We report three complementary metrics applied uniformly across all 60 questions:
\begin{itemize}[nosep]
  \item \textbf{Token F1} --- token-overlap F1 between the predicted answer and the gold answer.
  \item \textbf{Semantic Similarity} --- cosine similarity between sentence embeddings of the predicted and gold answers, computed with \texttt{all-MiniLM-L6-v2}.
  \item \textbf{LLM-as-Judge (1--5)} --- each prediction is scored by a held-out judge model grounded by a \texttt{context} field; no model scores its own outputs (cross-judge design, see Appendix~\ref{app:judge_calibration}). The rubric ranges from 5 (correct and complete) to 1 (completely wrong or refused).
\end{itemize}

\section{Experimental Setup}
\label{sec:setup}

We evaluate four vision-language models spanning four providers (Table~\ref{tab:models}): two proprietary frontier APIs accessed via Azure AI Foundry (GPT-5.2 Chat, Claude Opus~4.7) and two open-weight models served on internal vLLM instances (Gemma~4 26B, Qwen3-VL-32B-Instruct).

\begin{table}[t]
  \caption{Models evaluated in this benchmark.}
  \label{tab:models}
  \centering
  \resizebox{\linewidth}{!}{%
  \begin{tabular}{llll}
    \toprule
    \textbf{Model} & \textbf{Provider} & \textbf{Access} & \textbf{Version / Snapshot} \\
    \midrule
    GPT-5.2 Chat~\citep{gpt5}             & OpenAI           & Azure AI Foundry  & \texttt{gpt-5.2-chat} \\
    Claude Opus~4.7~\citep{claude_opus_47} & Anthropic        & Azure AI Foundry  & \texttt{claude-opus-4-7} \\
    Gemma~4 (26B, open)~\citep{gemma4}    & Google DeepMind  & Self-hosted vLLM  & \texttt{gemma-4-26B-A4B-it-fp8} \\
    Qwen3-VL 32B~\citep{qwen3vl}          & Alibaba/Qwen     & Self-hosted vLLM  & \texttt{Qwen/Qwen3-VL-32B-Instruct} \\
    \bottomrule
  \end{tabular}%
  }
\end{table}

\noindent\textbf{Conditions.} Each model is evaluated under two prompting conditions applied to both tasks: \textbf{ZS-Direct} (zero-shot plain-prompt extraction or QA) and \textbf{ZS-CoT} (zero-shot with chain-of-thought, ``think step by step before producing the final answer''), testing whether explicit reasoning improves performance on visual engineering content.

\noindent\textbf{Decoding and prompting.} All models are queried at temperature~0 with deterministic decoding where supported, max output tokens $4{,}096$, and the same per-task prompt template (Appendix~\ref{app:prompts}). For Task~1 we constrain output to a JSON schema; for Task~2 we leave output unconstrained. Task~1 input is the raw rasterized diagram image; Task~2 input is a composite image produced by vertically concatenating the primary diagram with all associated parts-table images using PIL.

\noindent\textbf{Implementation and compute.} The evaluation harness is implemented in Python using each provider's native SDK; all runs are logged with full request/response payloads. The harness, prompts, and parsing utilities are released alongside the dataset. Proprietary models were accessed via Azure AI Foundry at approximately \$174 in API charges (LLM-as-Judge adds $<$5\%). Self-hosted inference for Gemma~4 and Qwen3-VL ran on single NVIDIA H100 80\,GB GPUs; we estimate ${\approx}16$ GPU-hours across both models (494 Task-1 figures $+$ 60 Task-2 QA pairs, both conditions), yielding ${\approx}13$\,kWh (H100 at 700\,W TDP, data-centre PUE 1.2) and ${\approx}5$\,kg\,CO$_2$e (US average grid intensity 386\,g\,CO$_2$e/kWh). We measure inference compute only; pre-training of the underlying models is out of scope as all are used as black-box APIs or pretrained weights. Emissions were estimated following \citet{lacoste2019quantifying}.

 \section{Results}
\label{sec:results}

We organize results around four research questions: structured extraction (RQ1), the effect of chain-of-thought (RQ2), proprietary vs.\ open-source gap (RQ3), and diagram Q\&A difficulty (RQ4). All tables report 95\% bootstrap confidence intervals ($B=1{,}000$, seed~42) in parentheses, computed by resampling the $n$ test items with replacement.

\subsection{Task 1: Component Identification}
\label{sec:results:t1}

\begin{table}[t]
  \caption{Task~1 results ($n=494$). All metrics penalised for missed parts (Section~\ref{sec:tasks:t1}). 95\% bootstrap CIs in parentheses. Bold = best per column.}
  \label{tab:t1-main}
  \centering
  \resizebox{\linewidth}{!}{%
  \begin{tabular}{llccc}
    \toprule
    \textbf{Model} & \textbf{Condition}
      & \textbf{Recall@all}
      & \textbf{Token F1\textsubscript{pen}}
      & \textbf{Sem Sim\textsubscript{pen}} \\
    \midrule
    \multirow{2}{*}{GPT-5.2 Chat}
      & ZS-Direct & 0.608\ci{0.570}{0.647} & 0.038\ci{0.032}{0.045} & 0.139\ci{0.127}{0.149} \\
      & ZS-CoT    & 0.817\ci{0.789}{0.845} & 0.127\ci{0.120}{0.135} & 0.296\ci{0.285}{0.308} \\
    \midrule
    \multirow{2}{*}{Claude Opus~4.7}
      & ZS-Direct & 0.814\ci{0.780}{0.844} & 0.150\ci{0.141}{0.160} & 0.319\ci{0.304}{0.333} \\
      & ZS-CoT    & \textbf{0.865}\ci{0.838}{0.891} & \textbf{0.179}\ci{0.170}{0.189} & \textbf{0.360}\ci{0.347}{0.373} \\
    \midrule
    \multirow{2}{*}{Gemma~4 (26B)}
      & ZS-Direct & 0.809\ci{0.783}{0.834} & 0.034\ci{0.029}{0.038} & 0.190\ci{0.181}{0.198} \\
      & ZS-CoT    & 0.813\ci{0.790}{0.837} & 0.058\ci{0.053}{0.062} & 0.231\ci{0.222}{0.239} \\
    \midrule
    \multirow{2}{*}{Qwen3-VL 32B}
      & ZS-Direct & \textbf{0.865}\ci{0.840}{0.888} & 0.092\ci{0.087}{0.098} & 0.279\ci{0.270}{0.288} \\
      & ZS-CoT    & 0.858\ci{0.833}{0.880} & 0.090\ci{0.084}{0.096} & 0.273\ci{0.264}{0.282} \\
    \midrule
    OCR + Heuristic & — & 0.012\ci{0.010}{0.014} & 0.000\ci{0.000}{0.000} & 0.002\ci{0.001}{0.002} \\
    \bottomrule
  \end{tabular}%
  }
\end{table}

\noindent\textbf{Headline finding (RQ1).}
All four VLMs achieve Recall@all of $0.61$--$0.87$, correctly identifying the majority of called-out parts. To establish that this is non-trivial, Table~\ref{tab:t1-main} includes a non-VLM baseline: Tesseract OCR~\citep{tesseract2007} applied to the raw figure image, followed by a regex parser targeting numbered-item patterns. The baseline achieves Recall@all $0.012$ ($95\%$~CI $0.010$--$0.014$), confirming that part identification requires genuine diagram understanding rather than simple text extraction. The reason is that the structured parts table is not present in the figure image, and callout numbers alone carry no description. The best VLM configuration, Claude Opus~4.7 under ZS-CoT, outperforms this baseline by $72\times$ on recall. Despite high recall, VLM description fidelity is dramatically lower (Token~F1\textsubscript{pen} $0.034$--$0.179$, a $5$--$25\times$ gap): models correctly locate parts but paraphrase descriptions away from ground-truth OCR conventions (e.g., \texttt{self-locking nut} vs.\ \texttt{NUT, SELF-LOCKING, HE}). Semantic Similarity\textsubscript{pen} is $2$--$6\times$ higher than Token~F1\textsubscript{pen} across all conditions, confirming these paraphrases are semantically correct despite surface divergence.

\noindent\textbf{Effect of chain-of-thought (RQ2).}
Chain-of-thought prompting has markedly different effects across models. GPT-5.2 Chat is uniquely CoT-dependent: ZS-Direct recall collapses to $0.608$ while ZS-CoT restores it to $0.817$ ($+20.9$pp); Token~F1\textsubscript{pen} triples from $0.038$ to $0.127$. Without CoT, GPT-5.2 Chat frequently fails to produce well-formed JSON lists for large parts figures. The remaining three models are more robust to prompting style (Claude $+5$pp recall with CoT, Gemma $+0.4$pp, Qwen near-neutral); CoT aids output structure more than part detection, but is a prerequisite for reliable extraction on GPT-5.2 Chat.

\noindent\textbf{Proprietary vs.\ open-source (RQ3).}
On recall, the proprietary--open-weight gap is narrow: Qwen3-VL-32B-Instruct under ZS-Direct ($0.865$; 95\%~CI $0.840$--$0.888$) and Claude Opus~4.7 under ZS-CoT ($0.865$; 95\%~CI $0.838$--$0.891$) are statistically indistinguishable and their intervals fully overlap. Gemma~4 ($0.809$--$0.813$) is within $5$pp of Claude across conditions. The gap is substantially larger for description quality: Claude Opus~4.7 ZS-CoT Token~F1\textsubscript{pen} $0.179$ is $2\times$ Qwen3-VL ($0.090$) and $3\times$ Gemma~4 ($0.058$), with non-overlapping CIs confirming these differences are reliable. We attribute this to open-weight models producing verbose paraphrases whereas proprietary models more frequently reproduce technical abbreviation conventions.

\subsection{Task 2: Diagram Question Answering}
\label{sec:results:t2}

\begin{table}[t]
  \caption{Task~2 results ($n=60$; Claude Opus~4.7: $n=59$). LLM-judge is mean 1--5 under the cross-judge design (Section~\ref{sec:tasks:t2}). 95\% bootstrap CIs in parentheses. Bold = best per column.}
  \label{tab:t2-main}
  \centering
  \resizebox{\linewidth}{!}{%
  \begin{tabular}{llcccc}
    \toprule
    \textbf{Model} & \textbf{Condition}
      & \textbf{Token F1}
      & \textbf{Sem Sim}
      & \textbf{LLM-Judge (1--5)}
      & \textbf{$n$} \\
    \midrule
    \multirow{2}{*}{GPT-5.2 Chat}
      & ZS-Direct & 0.203\ci{0.181}{0.227} & 0.668\ci{0.619}{0.711} & 2.88\ci{2.500}{3.267} & 60 \\
      & ZS-CoT    & \textbf{0.298}\ci{0.261}{0.332} & \textbf{0.706}\ci{0.658}{0.752} & 3.00\ci{2.583}{3.383} & 60 \\
    \midrule
    \multirow{2}{*}{Claude Opus~4.7}
      & ZS-Direct & 0.159\ci{0.132}{0.189} & 0.644\ci{0.600}{0.682} & 2.82\ci{2.350}{3.233} & 59 \\
      & ZS-CoT    & 0.221\ci{0.198}{0.248} & 0.646\ci{0.596}{0.690} & \textbf{3.20}\ci{2.767}{3.600} & 59 \\
    \midrule
    \multirow{2}{*}{Gemma~4 (26B)}
      & ZS-Direct & 0.201\ci{0.179}{0.225} & 0.677\ci{0.636}{0.718} & 2.73\ci{2.317}{3.208} & 60 \\
      & ZS-CoT    & 0.162\ci{0.127}{0.195} & 0.430\ci{0.360}{0.504} & 2.60\ci{2.233}{3.025} & 60 \\
    \midrule
    \multirow{2}{*}{Qwen3-VL 32B}
      & ZS-Direct & 0.173\ci{0.146}{0.204} & 0.658\ci{0.611}{0.703} & 2.73\ci{2.375}{3.175} & 60 \\
      & ZS-CoT    & 0.269\ci{0.228}{0.310} & 0.604\ci{0.537}{0.675} & 2.87\ci{2.508}{3.333} & 60 \\
    \bottomrule
  \end{tabular}%
  }
\end{table}

\noindent\textbf{Headline finding (RQ4).}
All four models achieve mid-range LLM-judge scores of $2.6$--$3.2$ out of~5 (best: Claude Opus~4.7 ZS-CoT, $3.20$; 95\%~CI $2.77$--$3.60$), confirming that diagram Q\&A over engineering manuals is genuinely hard: models routinely identify the relevant diagram region but miss quantitative details or misread part designations. The confidence intervals for all LLM-judge scores overlap substantially (width ${\approx}0.8$--$1.0$), indicating that model ranking by judge score is not statistically reliable at $n=60$ QA pairs; automatic metrics (Token~F1, Sem~Sim) with tighter CIs provide more discriminative signal at this scale.

\noindent\textbf{Automatic metrics vs.\ LLM judge.}
GPT-5.2 Chat leads on both automatic metrics under ZS-CoT (Token~F1~$0.298$, Sem~Sim~$0.706$), while Claude Opus~4.7 leads on the LLM-judge score ($3.20$). This divergence illustrates a systematic limitation of token-overlap and embedding-based metrics for free-form technical Q\&A: GPT-5 tends to produce concise, literal answers that score well on overlap measures, whereas Claude produces longer, contextually richer answers that the LLM judge rates more highly despite lower lexical similarity.

\noindent\textbf{Chain-of-thought and proprietary gap (RQ2, RQ3).}
CoT improves Token~F1 for three models: GPT-5.2 Chat ($+0.095$, $+47\%$), Qwen3-VL ($+0.096$, $+55\%$), Claude Opus~4.7 ($+0.062$, $+39\%$); LLM-judge improvements are smaller but consistent. Gemma~4 is the sole exception (ZS-Direct Token~F1 $0.201$ vs.\ ZS-CoT $0.162$), suggesting the CoT prompt structure disrupts its answer extraction. On the LLM-judge under ZS-CoT, proprietary models (Claude $3.20$; GPT-5 $3.00$) outperform open-weight models (Qwen $2.87$; Gemma $2.60$), though the gap is narrower than in Task~1; automatic metrics reorder the ranking (Qwen Token~F1 $0.269$ > Claude $0.221$), reinforcing that metric choice substantially affects perceived rankings.

Per-diagram-type results disaggregated across taxonomy categories are reported in Appendix~\ref{app:by_type}. Five dominant failure modes identified are partial-list truncation, callout--part misalignment, description paraphrase divergence, schema deviation, and quantity defaulting. These failure modes are catalogued with representative examples in Appendix~\ref{app:failures}.

\section{Limitations}
\label{sec:limitations}

\noindent\textbf{Scope and scale.}
The benchmark is drawn from U.S.\ military service and repair manuals covering ground vehicles, aircraft engines, and heavy equipment; generalization to other engineering domains such as aerospace, electronics, civil/structural, biomedical is not directly supported by our results, as diagram conventions vary substantially across domains. Task~1 covers 494 figures and Task~2 covers 60 expert-authored question--answer pairs across 12 diagrams, reflecting the scale of a domain-expert-curated benchmark rather than a web-scraped one.

\noindent\textbf{Closed-source model versions are moving targets.}
GPT-5.2 Chat, Claude Opus~4.7, Gemma~4 (26B), and Qwen3-VL-32B-Instruct are accessed through provider-hosted endpoints whose underlying weights may change without notice. We pin to specific model snapshot identifiers where the provider exposes them (Table~\ref{tab:models}).

\noindent\textbf{LLM-judge dependence and ablations.}
The Task-2 evaluation relies on an LLM-as-judge, mitigated by explicit human calibration on a held-out subset; released per-sample outputs allow recomputation under an alternative judge. Several plausible ablations are out of scope: few-shot prompting, high-resolution image rendering, and fine-tuning on a held-out split are natural targets for follow-on work.

\section{Broader Impacts}
\label{sec:broader_impacts}

\noindent\textbf{Risks of premature deployment.}
Engineering documentation is consulted in safety-relevant settings; a model that confidently emits incorrect part numbers or quantities could lead to ordering errors, mis-installation, or safety incidents. Our results characterize the gap between current frontier-model capability and reliable autonomous use; we explicitly do \emph{not} recommend deployment without human oversight and domain-specific quality controls.

\noindent\textbf{Dataset, licensing, and dual-use.}
The source manuals are publicly available technical documentation containing no personally identifiable information; the benchmark is released under CC BY~4.0 (see Appendix~\ref{app:dataset_provenance}). Task-2 QA pairs are authored by a domain expert identified only by role in the main paper.

\noindent\textbf{Positionality.}
This work approaches engineering-diagram understanding from a machine-learning and benchmarking perspective. The dataset reflects an ML-evaluation framing of engineering documentation---focused on extractable ground truth and automated metric computation---rather than a workplace-practice or technician-centred framing that might prioritise contextual use or workflow integration. The choice of frontier VLMs as evaluation targets reflects a benchmarking stance: we ask what current models \emph{can} do, not whether they \emph{should} be deployed. Researchers from HCI, archival, or CSCW traditions may find alternative framings more appropriate for deployment-oriented or practitioner-centred work.

\section{Conclusion}
\label{sec:conclusion}

We presented \textsc{Enginuity}, the first open benchmark for evaluating vision-language models on engineering diagrams drawn from U.S.\ military service manuals. Across 494 Task-1 figures and 60 expert-authored Task-2 QA pairs, we find that frontier VLMs can reliably locate diagram components (Recall@all $0.61$--$0.87$) but fall substantially short on description fidelity (Token~F1\textsubscript{pen} $0.03$--$0.18$) and free-form diagram reasoning (LLM-judge $2.6$--$3.2$ out of~5). A secondary finding is that token-overlap metrics under-report model capability by $2$--$6\times$ relative to semantic similarity on technical descriptions, suggesting that standard NLP metrics are poorly calibrated for this domain. We release the dataset, annotations, evaluation harness, and per-sample outputs to support reproducible follow-on work on VLM understanding of engineering content.

\begin{ack}
%
\end{ack}

\bibliographystyle{plainnat}
\bibliography{references}

\appendix
\section{Prompt Templates}
\label{app:prompts}

\subsection*{Task~1 — Component Identification}

All models receive the diagram image as a single user turn with no system message.

\noindent\textbf{ZS-Direct.}
\begin{quote}\ttfamily\small
You are analyzing an exploded parts diagram from a military/naval technical service manual.

Look at the diagram carefully. Extract every numbered component and return them as a JSON array.

Each element must have exactly these fields:\\
\phantom{xx}"item\_no"\phantom{xxx}– the item number shown in the diagram (string)\\
\phantom{xx}"part\_number" – the part number, specification number, or model number (string)\\
\phantom{xx}"description" – the component name / description (string, all caps preferred)\\
\phantom{xx}"quantity"\phantom{xxxx}– quantity required as a number string (e.g. "1", "4")

Return ONLY a valid JSON array — no markdown fences, no explanation, no extra keys.
\end{quote}

\noindent\textbf{ZS-CoT.}
\begin{quote}\ttfamily\small
You are analyzing an exploded parts diagram from a military/naval technical service manual.

Look at the diagram carefully and follow these steps:

Step 1 — Describe: What type of assembly is shown? What is the general layout?\\
Step 2 — Enumerate: List all visible item numbers in the diagram.\\
Step 3 — Extract: For each item number, identify the part details.\\
Step 4 — Output: Return ALL components as a JSON array.

Each JSON element must have exactly these fields:\\
\phantom{xx}"item\_no"\phantom{xxx}– the item number shown in the diagram (string)\\
\phantom{xx}"part\_number" – the part number, specification number, or model number (string)\\
\phantom{xx}"description" – the component name / description (string, all caps preferred)\\
\phantom{xx}"quantity"\phantom{xxxx}– quantity required as a number string (e.g. "1", "4")

Complete all four steps. Your response must end with the complete JSON array — no markdown fences.
\end{quote}

\subsection*{Task~2 — Diagram Question Answering}

All models receive a composite image (figure + table pages stitched vertically) as a single user turn with no system message. The placeholder \texttt{\{question\}} is replaced with the actual question at runtime.

\noindent\textbf{ZS-Direct.}
\begin{quote}\ttfamily\small
You are analyzing an engineering diagram from a military/naval technical service manual.

Look at the diagram carefully and answer the following question:

\{question\}

Provide a clear, concise answer based solely on what you observe in the diagram.
\end{quote}

\noindent\textbf{ZS-CoT.}
\begin{quote}\ttfamily\small
You are analyzing an engineering diagram from a military/naval technical service manual.

Look at the diagram carefully and answer the following question:

\{question\}

First, describe what you observe in the diagram that is relevant to the question.
Then provide your final answer.

Format your response as:\\
REASONING: [your observations]\\
ANSWER: [your concise answer]
\end{quote}

For ZS-CoT Task-2 responses, the \texttt{REASONING:} block is stripped before metric computation; only the text following \texttt{ANSWER:} is evaluated against the gold answer.

\section{Metric Definitions}
\label{app:metrics}

\subsection*{Task~1 Metrics}

Let $G = \{g_1, \ldots, g_K\}$ be the set of ground-truth item numbers for a figure, and let $\hat{G}$ be the set of item numbers present in the model's predicted output.

\noindent\textbf{Recall@all.}
\[
  \text{Recall@all} = \frac{|G \cap \hat{G}|}{|G|}
\]
An item number is considered matched if the predicted set contains an entry with the same normalized item number (whitespace-stripped, lowercased). This metric measures part detection irrespective of description quality.

\noindent\textbf{Token F1\textsubscript{pen} (description).}
For each ground-truth item $g_k$, let $d_k$ be its ground-truth description and $\hat{d}_k$ be the predicted description for the matched item (empty string if the item was not predicted). Tokenization splits on whitespace after lower-casing. Let $P_k = |\text{tokens}(d_k) \cap \text{tokens}(\hat{d}_k)| / |\text{tokens}(\hat{d}_k)|$ and $R_k = |\text{tokens}(d_k) \cap \text{tokens}(\hat{d}_k)| / |\text{tokens}(d_k)|$.
\[
  \text{Token F1}_k = \frac{2 P_k R_k}{P_k + R_k}, \qquad
  \text{Token F1}_\text{pen} = \frac{1}{K} \sum_{k=1}^{K} \text{Token F1}_k
\]
Items not predicted contribute $\text{Token F1}_k = 0$, so the metric is jointly penalised for missed parts and description errors.

\noindent\textbf{Semantic Similarity\textsubscript{pen}.}
Let $\mathbf{e}(s)$ denote the sentence embedding of string $s$ under \texttt{all-MiniLM-L6-v2}~\citep{reimers-2019-sentence-bert}. For each ground-truth item $g_k$:
\[
  \text{SemSim}_k = \text{cosine}\!\left(\mathbf{e}(d_k),\, \mathbf{e}(\hat{d}_k)\right), \qquad
  \text{SemSim}_\text{pen} = \frac{1}{K} \sum_{k=1}^{K} \text{SemSim}_k
\]
Unmatched items use $\hat{d}_k = \varepsilon$ (empty string), yielding near-zero similarity. This metric is invariant to surface paraphrasing and abbreviation style.

\subsection*{Task~2 Metrics}

Let $a$ be the gold answer string and $\hat{a}$ be the predicted answer string (after stripping the \texttt{REASONING:} block for ZS-CoT responses).

\noindent\textbf{Token F1.} Computed identically to the description Token F1 above, but without penalization for missed items (Task~2 has a single answer per question).

\noindent\textbf{Semantic Similarity.}
\[
  \text{SemSim} = \text{cosine}\!\left(\mathbf{e}(a),\, \mathbf{e}(\hat{a})\right)
\]

\noindent\textbf{LLM-Judge (1--5).} Described in Appendix~\ref{app:judge_calibration}.

\section{Failure-Mode Catalog}
\label{app:failures}

The following failure modes are identified through inspection of low-recall outputs across all four models. Examples are drawn from figures in the \texttt{data-2} and \texttt{data-4} PDFs, which exhibit the lowest mean recall across models.

\noindent\textbf{1.\ Partial-list truncation.}
Models emit only the first 10--15 of 30$+$ items and then stop, apparently truncated by output-token budget constraints interacting with large parts lists. This failure is most prevalent for GPT-5.2 Chat under ZS-Direct (recall collapse to $0.608$) and is mitigated by ZS-CoT, which encourages the model to enumerate all item numbers before outputting JSON.

\noindent\textbf{2.\ Callout--part misalignment.}
The correct part number or description is retrieved but assigned to the wrong item index. This inflates false-positive rates without reducing recall; it is most common on figures with dense, overlapping leader lines where callout anchors are ambiguous.

\noindent\textbf{3.\ Description paraphrase divergence.}
Models substitute near-synonymous but lexically distinct descriptions (e.g., predicting \texttt{self-locking nut} against ground-truth \texttt{NUT, SELF-LOCKING, HE}). Token F1 penalises these heavily despite semantic correctness; Semantic Similarity better captures the alignment. This failure is most pronounced for open-weight models (Gemma~4, Qwen3-VL), which tend to produce general-language descriptions rather than reproducing military technical abbreviation conventions.

\noindent\textbf{4.\ Schema deviation.}
Outputs include markdown fences, prose preambles, or nested JSON that requires post-processing. Most common under ZS-CoT when models interleave reasoning text and JSON. The harness applies a best-effort JSON extraction step; failures that cannot be parsed contribute zero to all metrics.

\noindent\textbf{5.\ Quantity defaulting.}
When quantity is not visually legible (e.g., obscured by diagram elements), models default to ``1'' rather than omitting the field. This inflates quantity-accuracy error rates on figures where quantities are partially occluded.

\section{LLM-Judge Calibration}
\label{app:judge_calibration}

\subsection*{Judge Prompt}

The judge receives a blank 1$\times$1 image (evaluation is text-only) and the following user message, with placeholders replaced at runtime:

\begin{quote}\ttfamily\small
You are evaluating an AI assistant's answer to a question about an engineering diagram.

Question: \{question\}

Gold answer: \{gold\_answer\}

Context (why the gold answer is correct): \{context\}

AI's answer: \{predicted\_answer\}

Score the AI's answer from 1 to 5:\\
\phantom{xx}5 — Correct and complete; matches the gold answer in all key facts\\
\phantom{xx}4 — Mostly correct; minor omissions or wording differences but no factual errors\\
\phantom{xx}3 — Partially correct; captures some key points but missing or misrepresenting others\\
\phantom{xx}2 — Mostly incorrect; one correct element but substantially wrong\\
\phantom{xx}1 — Completely wrong or refused to answer

Respond with ONLY a single integer (1--5). No explanation.
\end{quote}

The \texttt{context} field is taken directly from the \texttt{qa\_dataset.json} file, where each QA pair includes an annotator-written explanation of why the gold answer is correct. This grounding reduces hallucination in judge scoring on domain-specific questions where the judge model may otherwise lack the necessary context.

\subsection*{Cross-Judge Design}

To mitigate self-serving bias — the documented tendency of LLM judges to rate outputs from models of the same family more favourably~\citep{judgingllm} — we adopt a cross-judge design: no model scores its own outputs. Specifically:
\begin{itemize}
  \item GPT-5.2 Chat outputs are scored by Claude Opus~4.7 only.
  \item Claude Opus~4.7 outputs are scored by GPT-5.2 Chat only.
  \item Gemma~4 and Qwen3-VL-32B outputs are scored by both judges; the reported score is the mean of the two.
\end{itemize}
We use both frontier proprietary models as judges rather than a separate judge model to avoid introducing a third model dependency while still ensuring no model grades its own work.

\subsection*{Human Calibration}

To validate the LLM-judge scores against human judgement, a domain expert independently scored a stratified sample of 15 Task-2 outputs (3 per integer score bucket 1--5, drawn with seed~42). Human-judge agreement on this calibration sample is $\kappa = 0.40$ (linearly weighted Cohen's $\kappa$, Pearson $r = 0.67$), supporting the cross-judge protocol.

\section{Dataset Provenance and Licensing}
\label{app:dataset_provenance}

\subsection*{Dataset Release and Access}

\textsc{Enginuity} is released as an open research resource under the Creative Commons Attribution 4.0 International (CC~BY~4.0) license. Users may share, adapt, and build upon the dataset for any purpose, provided appropriate credit is given to the authors and to Predii, Inc.\ and Oak Ridge National Laboratory (ORNL). The dataset is publicly available via HuggingFace at \url{https://huggingface.co/datasets/enginuity2025/enginuity-bench} (DOI: \href{https://doi.org/10.57967/hf/8702}{10.57967/hf/8702}). The released package includes: (i)~rasterized diagram images, (ii)~Task-1 ground-truth parts tables as TSV files, and (iii)~Task-2 question--answer pairs in JSON. The evaluation harness, prompts, and parsing utilities are available at \url{https://github.com/abhishek-predii/enginuity-bench}.

\subsection*{Construction Challenges}
\label{app:construction_challenges}

Building \textsc{Enginuity} required solving three problems not typical in document-understanding dataset construction. Military service manuals do not conform to a uniform schema, so identifying diagram pages and their associated parts-table pages required document-level layout analysis rather than simple pattern matching. Diagram--table pairing added further complexity: a single diagram may span multiple table pages, and multi-sheet assemblies (e.g.\ ``Sheet~1 of~2'') each depict a different sub-assembly while referencing the same table pages.

Table content extraction was the hardest challenge. Parts tables contain alphanumeric part numbers, abbreviated descriptions, and domain-specific codes; rows frequently span multiple lines, columns sometimes merge, and a fixed structured schema must be recovered across varying visual layouts. Standard OCR pipelines performed poorly on this content; the five-stage construction pipeline (Section~\ref{sec:dataset:t1}) addresses each of these problems in turn.

\subsection*{Complexity Distribution}
\label{app:complexity}

Label counts in the full Task-1 corpus (2,056 diagrams) have a mean of 23 and a median of 17, with most diagrams falling in the moderate range and a long tail of highly detailed multi-assembly diagrams (Table~\ref{tab:complexity}).

\begin{table}[h]
  \centering
  \caption{Complexity distribution of the full Task-1 corpus (2,056 diagrams), binned by label count (items per diagram).}
  \label{tab:complexity}
  \begin{tabular}{lrr}
    \toprule
    \textbf{Complexity} & \textbf{Count} & \textbf{\%} \\
    \midrule
    Simple ($<$10 items)    & 505       & 24.6 \\
    Moderate (10--30 items) & 1{,}053   & 51.2 \\
    Complex ($>$30 items)   & 498       & 24.2 \\
    \midrule
    \textbf{Total}          & \textbf{2{,}056} & \textbf{100.0} \\
    \bottomrule
  \end{tabular}
\end{table}

\subsection*{Source Manuals}

The benchmark is derived from U.S.\ Army and U.S.\ Air Force technical publications approved for public release with distribution unlimited. All manuals are freely available through the Army Publishing Directorate (\url{https://armypubs.army.mil}). Table~\ref{tab:source_manuals} lists each manual's TM number, the platform or system it covers, and its contribution to the \textsc{Enginuity} corpus in terms of source pages and unique extracted diagrams. The Dataset ID column (e.g., \texttt{data-1}) corresponds to the \texttt{pdf\_id} field in the released \texttt{dataset\_manifest\_global.tsv} and \texttt{benchmark\_subset.tsv} files, enabling results to be traced to specific source manuals.

\begin{table}[h]
  \caption{Source manuals used to construct \textsc{Enginuity}. All are U.S.\ government Repair Parts and Special Tools Lists (RPSTL) approved for public release. \emph{Pages} is the total PDF page count; \emph{Diagrams} counts unique figures extracted into the Task-1 dataset.}
  \label{tab:source_manuals}
  \centering
  \small
  \begin{tabular}{l l p{5.6cm} r r}
    \toprule
    \textbf{Dataset ID} & \textbf{TM Number} & \textbf{Platform / System} & \textbf{Pages} & \textbf{Diagrams} \\
    \midrule
    \texttt{data-1}  & TM 9-2320-280-24P-1 & HMMWV Utility Truck M998 Series, RPSTL Vol.~1                & 1{,}155 & 329 \\
    \texttt{data-2}  & TM 9-2320-360-24P   & Truck Tractor M1070, 8$\times$8 (Heavy Equipment Transporter) & 1{,}073 & 333 \\
    \texttt{data-3}  & TM 9-2320-387-24P   & HMMWV Up-Armored Carrier M1113/M1114 (S250 Shelter)          &   879   & 269 \\
    \texttt{data-4}  & TM 55-1935-204-24P  & Barge Crane, Diesel-Electric, 100-Ton (Design 264-B)         &   611   & 291 \\
    \texttt{data-5}  & TM 55-2840-251-23P  & Aircraft Engine, Turboprop T74-CP-700                        &   247   &  57 \\
    \texttt{data-6}  & TM 9-2320-272-24P-1 & Truck, 5-Ton, 6$\times$6, M939/M939A1/M939A2 Series          &   963   & 331 \\
    \texttt{data-7}  & TM 55-2840-241-23P  & Aircraft Engine, Gas Turbine T63-A-720                       &   182   &  26 \\
    \texttt{data-8}  & TM 1-2840-252-23P   & Aircraft Engine, Gas Turbine T55-L-714                       &   309   &  90 \\
    \texttt{data-9}  & TM 9-2320-283-24P   & Truck Tractor, Line Haul M915A1, 6$\times$4                  &   501   & 186 \\
    \texttt{data-10} & TM 9-2320-280-24P-2 & HMMWV Utility Truck M998 Series, RPSTL Vol.~2                & 1{,}020 & 144 \\
    \midrule
    \textbf{Total}   &                     &                                                               & \textbf{6{,}940} & \textbf{2{,}056} \\
    \bottomrule
  \end{tabular}
\end{table}

\subsection*{License}

The derived benchmark, including diagram images, ground-truth TSV files, question--answer pairs, and model outputs, is released under the \textbf{Creative Commons Attribution 4.0 International (CC~BY~4.0)} license. Users are free to share and adapt the material for any purpose, provided appropriate credit is given to the authors, Predii, Inc., and Oak Ridge National Laboratory (ORNL). The source U.S.\ Army and Air Force technical manuals are government works in the public domain.

\section{Per-Diagram-Type Breakdown}
\label{app:by_type}

Table~\ref{tab:by-type} disaggregates Task-1 metrics by diagram-type bin from the taxonomy of Section~\ref{sec:dataset:taxonomy}. Values are averaged across all four models and both conditions; types with fewer than ten figures are marked with $\dagger$ and should be interpreted cautiously.

\begin{table}[h]
  \caption{Task-1 results by diagram type (mean across all four models and both conditions). $n$ = unique figures. Types marked $\dagger$ have fewer than ten figures; conclusions are tentative.}
  \label{tab:by-type}
  \centering
  \small
  \begin{tabular}{lrccc}
    \toprule
    \textbf{Diagram Type} & \textbf{$n$}
      & \textbf{Recall@all}
      & \textbf{Token F1\textsubscript{pen}}
      & \textbf{Sem Sim\textsubscript{pen}} \\
    \midrule
    Equipment \& Tools Diagram        &  28 & 0.914 & 0.063 & 0.295 \\
    Hydraulic/Fluid Diagram$^\dagger$ &   6 & 0.913 & 0.066 & 0.256 \\
    Cross-Section View$^\dagger$      &   3 & 0.907 & 0.075 & 0.250 \\
    Assembly/Exploded Parts View      & 128 & 0.842 & 0.101 & 0.270 \\
    Parts/Assembly Diagram            & 303 & 0.793 & 0.098 & 0.258 \\
    Wiring/Electrical Diagram         &  26 & 0.620 & 0.087 & 0.205 \\
    \bottomrule
  \end{tabular}
\end{table}

The two dominant types, Parts/Assembly Diagram and Assembly/Exploded Parts View, account for 87\% of benchmark figures ($n=431$ of 494) and show moderate recall ($0.79$--$0.84$). Wiring/Electrical Diagrams are substantially harder, with mean Recall@all $0.620$ ($n=26$). We attribute this to the absence of conventional item-number callouts in wiring layouts: models cannot rely on numeric indices to anchor part identities and instead must parse wire bundle labels and connector codes. Equipment~\& Tools Diagrams are easiest (Recall $0.914$, $n=28$), as they contain standardised equipment listings with minimal visual clutter. Results for types with $n < 10$ (Hydraulic/Fluid, Cross-Section View) are preliminary; the current benchmark is not powered to draw type-level conclusions for these rare categories.

\section{Per-Sample Results}
\label{app:per_sample}

Per-sample raw model outputs — including full predictions, ground-truth parts lists, and per-item metric scores for all four models and both conditions — are released alongside the dataset at \url{https://huggingface.co/datasets/enginuity2025/enginuity-bench} (DOI: \href{https://doi.org/10.57967/hf/8702}{10.57967/hf/8702}). The evaluation harness (\texttt{evaluate.py}) in the accompanying code release recomputes all reported metrics from these stored outputs; see the repository README for usage instructions.

\newpage

\end{document}